\documentclass[letterpaper]{article} % DO NOT CHANGE THIS
\usepackage{aaai25}  % DO NOT CHANGE THIS
\usepackage{times}  % DO NOT CHANGE THIS
\usepackage{helvet}  % DO NOT CHANGE THIS
\usepackage{courier}  % DO NOT CHANGE THIS
\usepackage[hyphens]{url}  % DO NOT CHANGE THIS
\usepackage{graphicx} % DO NOT CHANGE THIS
\usepackage{natbib}  % DO NOT CHANGE THIS AND DO NOT ADD ANY OPTIONS TO IT
\usepackage{caption} % DO NOT CHANGE THIS AND DO NOT ADD ANY OPTIONS TO IT
\usepackage{algorithm}
\usepackage{algorithmic}
\usepackage{alltt}
\usepackage{amsmath,amsfonts}
\usepackage{booktabs}
\usepackage{multirow}
\usepackage{varwidth}
\usepackage{comment}
\usepackage{xcolor}
\definecolor{customblue}{RGB}{39, 161, 185}
\definecolor{customred}{RGB}{68, 114, 196}
\usepackage{bbding}
\usepackage{graphicx} % 为\resizebox命令加载宏包
\usepackage{array}    % 为使用\newcolumntype加载宏包
\usepackage{colortbl} % 为使用\cellcolor命令加载宏包
\usepackage{subcaption} % 为使用子表格加载宏包
\usepackage{makecell}
\usepackage{enumitem}
\usepackage{bibentry}
\usepackage{newfloat}
\usepackage{listings}
\usepackage{subcaption}
\usepackage{float}
\usepackage{bm}
\usepackage{amssymb}

\usepackage{newfloat}
\usepackage{listings}
\DeclareCaptionStyle{ruled}{labelfont=normalfont,labelsep=colon,strut=off} % DO NOT CHANGE THIS
\floatstyle{ruled}
\newfloat{listing}{tb}{lst}{}
\floatname{listing}{Listing}
\nocopyright
\title{ComKD-CLIP: Comprehensive Knowledge Distillation for Contrastive Language-Image Pre-traning Model}
\author{
    \
    Yifan Chen\textsuperscript{\rm 1}\textsuperscript{\rm 3}\equalcontrib,
    Xiaozhen Qiao\textsuperscript{\rm 2}\textsuperscript{\rm 3}\equalcontrib,
    Zhe Sun\textsuperscript{\rm 1}\textsuperscript{\rm 3}\thanks{Corresponding author: sunzhe@nwpu.edu.cn} \& 
    Xuelong Li\textsuperscript{\rm 1}\textsuperscript{\rm 3}\thanks{Corresponding author: li@nwpu.edu.cn}
}

\affiliations{
    \textsuperscript{\rm 1}School of Artificial Intelligence, OPtics and ElectroNics (iOPEN), Northwestern Polytechnical University\\
    \textsuperscript{\rm 2}School of Information Science and Technology, University of Science and Technology of China\\
    \textsuperscript{\rm 3}Institute of Artificial Intelligence (TeleAI), China Telecom\\
    
}

\usepackage{bibentry}
\begin{document}

\maketitle
\begin{abstract}
Contrastive Language-Image Pre-training (CLIP) models
excel in integrating semantic information between images and texts through contrastive learning techniques. It has achieved remarkable performance in various multimodal tasks. However, the deployment of large CLIP models is hindered in resource-limited environments, while smaller models frequently fail to meet the performance benchmarks required for practical applications. In this paper, we propose a novel approach, ComKD-CLIP: Comprehensive Knowledge Distillation for Contrastive Language-Image Pre-traning Model, which aims to comprehensively distill the knowledge from a large teacher CLIP model into a smaller student model, ensuring comparable performance with significantly reduced parameters. ComKD-CLIP is composed of two key mechanisms: Image Feature Alignment (IFAlign) and Educational Attention (EduAttention). IFAlign makes the image features extracted by the student model closely match those extracted by the teacher model, enabling the student to learn
teacher’s knowledge of extracting image features. EduAttention explores the  cross-relationship between text features extracted by the teacher model and image features extracted by the student model, enabling the student model to learn how the
teacher model integrates text-image features.
In addition, ComKD-CLIP can refine the knowledge distilled from IFAlign and EduAttention by leveraging the text-image feature fusion results of the teacher model, ensuring the student model accurately absorbs the teacher's knowledge. Extensive experiments conducted on 11 datasets have demonstrated the superiority of the proposed method. 
\end{abstract}

\section{Introduction}

\begin{figure}[!h]
 \centering
\includegraphics[width=\columnwidth]{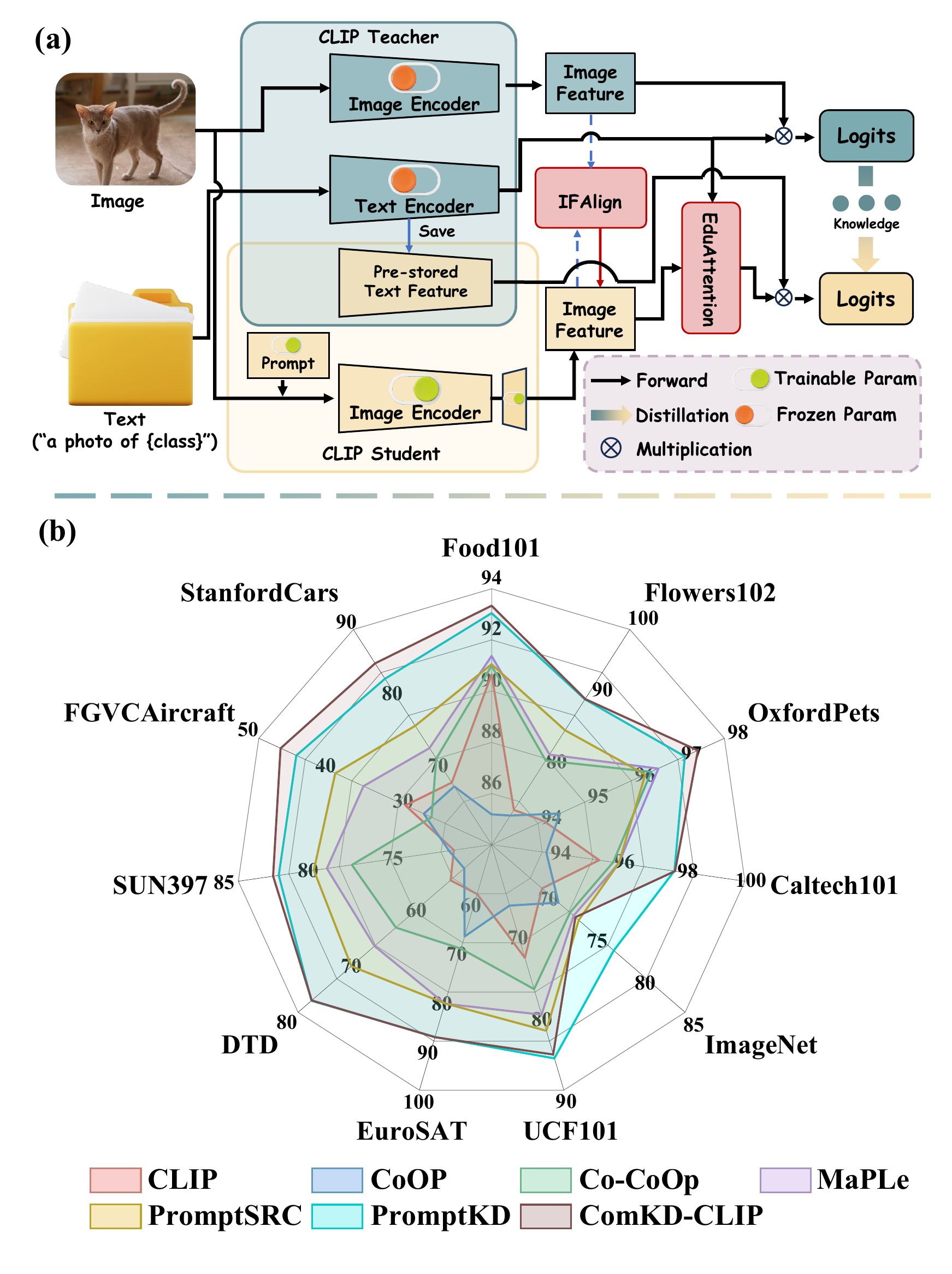}
\caption{(a) Schematic of proposed ComKD-CLIP framework. (b) Comparison of the Harmonic Mean (HM) for generalization from base to novel categories. All methods utilize the ViT-B/16 image encoder from the pretrained CLIP model. ComKD-CLIP is proud to achieve the \textbf{best performance} in 8 out of 11 diverse recognition datasets.}
\label{fig:1}  
\vspace{-10pt}
\end{figure}

Contrastive Language-Image Pre-training (CLIP) \cite{radford2021learning} has emerged as a  prominent pre-trained Visual Language Model (VLM) due to its ability to effectively integrate the semantic information between images and texts through contrastive learning techniques. This ability has enabled CLIP to demonstrate strong performance across various multimodal tasks  such as image recognition, visual question answering, and image description generation \cite{li2023scaling, wei2022mvp, singh2022flava}. However, large CLIP models cannot be deployed in resource-constrained environments and the performance of smaller CLIP models often falls short of application requirements. These limitations significantly restrict the practical applicability of CLIP models.

Knowledge Distillation (KD) \cite{hinton2015distilling} has been introduced 
into CLIP to address these issues. CLIP-TD \cite{wang2022clip} distills knowledge from both CLIP’s vision and language branches into existing architectures for VL tasks. TinyCLIP \cite{wu2023tinyclip} proposes affinity mimicking and weight inheritance to improve the performance of small models by leveraging large-scale models. CLIPPING \cite{pei2023clipping} proposes a new layer-wise alignment with the student as the base, which enables the student to fully absorb the knowledge of the teacher. PromptKD \cite{li2024promptkd} uses unlabeled domain data to perform prompt-based knowledge distillation for CLIP, which can greatly improve the performance of small model. CLIP-KD \cite{yang2024clip} proposes several distillation strategies to examine the effectiveness of CLIP-KD. However, these methods distill the knowledge of the teacher model based on the results of text-image feature fusion, ignoring the knowledge embedded within the fusion process. This oversight substantially hinders the student model to absorb the teacher model's knowledge.

To ensure that the student model can fully absorb the knowledge of the teacher model, we endeavor to distill the knowledge embedded in the feature fusion process of teacher model and refine the distilled knowledge leveraging the feature fusion results of teacher model. Accordingly, we propose ComKD-CLIP: Comprehensive Knowledge Distillation for Contrastive Language-Image Pre-traning Model. ComKD-CLIP is composed of two key modules: Image Feature Alignment (IFAlign) and Educational Attention (EduAttention). During the feature fusion stage: IFAlign ensures that the image features extracted by the student model closely match those extracted by the teacher model. This alignment enable the student to absorb the teacher's knowledge of how to extract image feature. Concurrently, EduAttention explores the cross-relationship between the text features extracted by the teacher model and the image features extracted by the student model. Through this strategy, EduAttention enables the student model to comprehend and emulate the teacher model's abilities for integrating image and text features, thus enriching its own mutlimodal understanding. Furthermore, to make the student model accurately absorb the knowledge of the teacher model, ComKD-CLIP refines the knowledge distilled from ImgAlign and EduAttention leveraging the feature fusion results of the teacher model. The schematic of proposed ComKD-CLIP is shown as Fig. \ref{fig:1}  (a). We also compare the performance of the proposed method with some state-of-the-art methods on 11 datasets. The experimental results are shown in Fig. \ref{fig:1} (b). It is worth mentioning that the proposed method has best performance in 8 out of 11 diverse recognition datasets.

The main contributions of the proposed method can be summarized as follows:

 \begin{itemize}
        \item We propose an IFAlign module that enables student model to absorb the teacher model's knowledge on how to extract image features during the process of text-image feature fusion.
        \item We propose an EduAttention module that enables student model to absorb the teacher model's knowledge on how to integrate the text-image features during the process of text-image feature fusion.
        \item We refine the knowledge distilled from IFAlign and EduAttention leveraging the feature fusion results of the teacher model, which can prompt the student model to accurately absorb the teacher model's knowledge. Extensive experiments conducted on 11 datasets have demonstrated the superiority of the proposed ComKD-CLIP.

        \end{itemize}

\section{Related Work}

\subsection{Contrastive Language-Image Pre-training (CLIP)}

CLIP can simultaneously comprehend and fuse the text-image data, achieving outstanding performance in multimodal tasks \cite{gao2022pyramidclip, zhu2023not,li2024coleclip}.
SLIP \cite{mu2022slip} extends CLIP's capabilities by integrating it with self-supervised learning, facilitating its application within the domain of multi-task learning.
MaskCLIP \cite{dong2023maskclip} innovates further by introducing masked self-distillation, a technique that distills the representation of a complete image onto the predicted representation of its masked counterpart. This approach significantly enhances CLIP's performance. AttCLIP \cite{yang2023attentive} incorporates attention mechanisms into CLIP, enabling the model to selectively focus on tokens that exhibit a high degree of correlation with the corresponding textual information. This method not only facilitates efficient multi-view learning but also economizes on training time. CLIP-Decoder \cite{ali2023clip} enriches the multi-modal representation learning of CLIP through the integration of distinct encoders for text and images, leading to significant advancements in multi-label classification tasks. MoPE-CLIP \cite{lin2024mope} proposes a novel module-wise pruning error metric, allowing for the effective leveraging of teacher model knowledge. This provides a unified solution for the pre-training phase of CLIP models. Collectively, these advancements underscore CLIP's prowess in multimodal tasks. However, the deployment of large CLIP models remains constrained in resource-limited environments, whereas small models often fail to meet the requisite benchmarks for practical utility. Consequently, the primary challenge lies in compressing CLIP models without compromising their performance, thus facilitating their widespread applicability across diverse computational landscapes.

\subsection{Knowledge Distillation (KD)}

KD  aims to enable the small student model to absorb knowledge from the large teacher model, thus achieving comparable performance to the large model.
It has achieved remarkable success in numerous vision tasks, including image segmentation \cite{liu2019structured, yang2022cross}, object detection \cite{jia2024mssd, wang2024crosskd}, and
pose estimation \cite{li2021online}. Recently, many researchers have endeavored to introduce KD into CLIP,  motivated by the pressing need to surmount the operational challenges faced by large CLIP models in resource-limited environments, alongside the performance shortfalls exhibited by smaller models in practical applications.\cite{laroudie2023improving}. 
CLIPPING \cite{pei2023clipping} introduces a novel layer-wise alignment strategy that takes the student model as the foundation, enabling the student model to effectively absorb the knowledge from the teacher model. PromptKD \cite{li2024promptkd}, in a distinct approach, capitalizes on unlabeled domain data to facilitate prompt-based knowledge distillation within the CLIP paradigm, significantly bolstering the performance of smaller CLIP models. TinyCLIP \cite{wu2023tinyclip}, similarly targeting CLIP distillation, achieves commendable results through the innovative application of affinity mimicking and weight inheritance techniques. However, prior studies predominantly concentrate on  distilling the knowledge of teacher model based on feature fusion results, overlooking the intricate knowledge encapsulated within the feature fusion process. In stark contrast to existing distillation methodologies, the proposed method uniquely distills the knowledge inherent to the text-image feature fusion process within teacher models. In addition, by refining the distilled knowledge leveraging the feature fusion results, ComKD-CLIP enables the student model to accurately absorb the nuanced knowledge from teacher model, thereby precipitating a marked improvement in their performance.

\section{Approach}

\subsection{Preliminaries}

\textbf{CLIP} is one of the most commonly used VLMs, comprising independent image and text encoder branches. It aligns and fuses images with texts to learn a joint multimodal embedding space. In the image encoding branch, a labeled visual recognition dataset $D=\{x_{j}, y_{j}\}_{j=1}^{M}$ is used as input. Each image $x$ from dataset $D$ is processed by the image encoder $f_{I}$ to obtain normalized image features $u=f_{I}(x)/||f_{I}(x)||_{2} \in \mathbb{R}^{d}$. Corresponding to the image recognition dataset $D$ has $N$ class names $\text{c}=\{c_{i}\}_{i=1}^{N}$. In the text encoder branch, the input data is text descriptions $t_{i}$ generated from the template “a photo of a \{$c_{i}$\}”. Each $t_{i}$, after being encoded by the text encoder, yields normalized text features $w_{i}=f_{T}(t_{i})/||f_{T}(t_{i})||_{2} \in \mathbb{R}^{d}$, where $d$ is the dimension of the text features. The collection of all text features $W=[w_{1}, w_{2}, \ldots, w_{N}] \in \mathbb{R}^{N \times d}$ serves as the classification weight matrix. Based on this data, the classification output probability can be calculated as follows:

\begin{equation}
    p(y|x) = \frac{\text{exp}(u w_{y}^{\mathsf{T}}/\tau)}{\sum_{i=1}^{N}\text{exp}(u w_{i}^{\mathsf{T}}/\tau)},
\label{equation:output_prob}
\end{equation}
where $uw^{\mathsf{T}}$ represents the output logit and $\tau$ is the temperature parameter.

\noindent \textbf{KD} is originally proposed by Hinton et al. \cite{hinton2015distilling}, it transfers knowledge from a large, pre-trained teacher model to a smaller, lightweight student model. It can help the student to absorb the teacher's knowledge for efficient deployment. This process employs the KL divergence loss to align the feature distributions of both models. The KL divergence loss is defined as follows:

\begin{equation}
L_{kd}(q^{t}, q^{s}, \tau) = \tau^{2} KL(\sigma (q^{t}/\tau),\sigma (q^{s}/\tau)),
\label{equation:kd}
\end{equation}
where $q^{t}$ and $q^{s}$ represent the logits predicted by the teacher and student models, respectively. $\sigma(\cdot)$  represents the softmax function, and $\tau$ is the temperature parameter~\cite{hinton2015distilling,li2023curriculum}, which adjusts the smoothness of the probability distribution.

\begin{figure*}[!h]
 \centering
\includegraphics[width=0.95\linewidth]{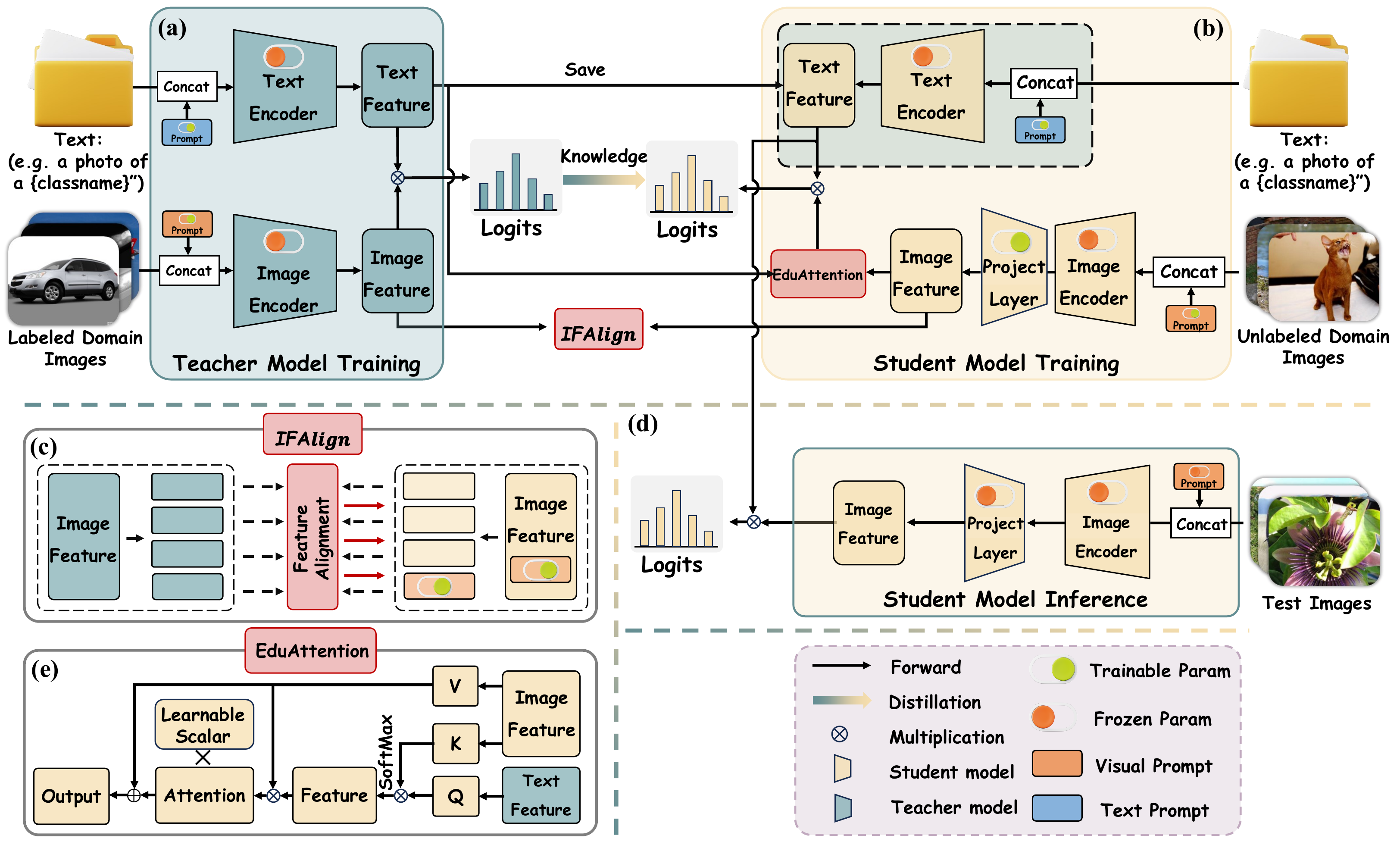}
\caption{Overview of our proposed ComKD-CLIP framework. (a) Utilization of the cue learning method with a well-trained large teacher CLIP model; (b) A smaller student CLIP model, which is trained with learnable cues and reuses the text features from the teacher model, requiring training only for the image encoder branch; (c) The schematic of IFAlign module; (d) The inference process within the trained student model, where the text encoder branch reuses the text features of the teacher model; (e) The schematic of EduAttention module.}
\label{fig:2}  
\vspace{-10pt}
\end{figure*}

\subsection{Pipline}

As illustrated in Fig.~\ref{fig:2}, our proposed ComKD-CLIP framework comprises two principal stages: the pretraining of the large CLIP teacher model and the subsequent training of a small CLIP student model. In the initial phase,  as delineated in Fig. ~\ref{fig:2}(a), the large CLIP teacher model is pretrained on a  labeled domain dataset, $D_{labeled}=\{x_{i}, y_{i}\}_{i=1}^{M}$, to enhance its performance,  aligning with contemporary methodologies such as PromptSRC~\cite{khattak2023self} and PromptKD~\cite{li2024promptkd}. Innovatively, we incorporate learnable prompts into both the image and text encoder branches of the teacher model via a concatenation strategy. The image and text data from the labeled domain dataset are processed through the image encoder $f_{V}^{t}$ and text encoder $f_{T}^{t}$, respectively, producing image features $u^p_t\in \mathbb{R}^{d}$ and text features $w^p_t\in \mathbb{R}^{d}$. The ultimate output logits $q^{t}$ is calculated by Eq.~\ref{equation:output_prob}. The training of the teacher model entails minimizing the cross-entropy loss between the predicted probability distributions and the true labels, thereby optimizing the model's parameters. This rigorous pretraining phase ensures the teacher model acquires robust knowledge that can be effectively distilled to the student model during the latter stage of our framework.

As depicted in Fig.~\ref{fig:2}(b), the student CLIP model directly capitalizes on the pre-trained text features from the teacher model, thereby significantly curtailing the training costs with the text encoder branch. Simultaneously, a lightweight CLIP image encoder branch is engineered within the student model to decrease resource costs while maintaining competitive performance. During the processing of input data from the unlabeled domain dataset $D_{unlabeled}$ by the student model's image encode, we incorporate the IFAlign module. This module serves to align the student model's image features $u^p_s\in \mathbb{R}^{d}$ with the teacher model’s image features $u^p_t\in \mathbb{R}^{d}$, thereby facilitating the student model to absorb the knowledge of how the teacher model extracts salient image features. Subsequent to the feature alignment, the EduAttention module is introduced to explore the cross-relationship between the image features extracted by student model and text features provided by teacher model. This exploration enables the student model to learn the nuanced strategies employed by the teacher model for integrating text-image feature. In addition, we employ the KL divergence to minimize the discrepancy between the logits produced by the teacher and student models. This optimization ensures that the knowledge distilled by the student model is refined and closely mirrors that of the teacher, thus enhancing the student model to accurately absorb the knowledge of teacher model. Finally, the inference process of trained student model is shown in Fig.~\ref{fig:2}(d).

% 如图1所示， 我们提出的对齐蒸馏框架包括两个部分，分别是大CLIP老师模型的预训练以及通过对齐蒸馏来训练轻量级的CLIP学生模型。如图1（a）所示，使用带有标签的域数据来训练大CLIP老师模型，为更好的释放老师模型的工作性能，就像现有的工作PromptSRC以及PromptKD一样，我们在视觉编码分支以及文本编码分支上面均通过“Concate”的方式添加了可学习的提示。将标签的域数据集$D_{labeled}=\{x_{i}, y_{i}\}_{i=1}^{M}$以及设定的类名称投喂进老师模型，视觉数据以及文本数据分别经过视觉编码器$f_{V}^{t}$以及文本编码器$f_{T}^{t}$得到视觉特征$u^p_t\in \mathbb{R}^{d}$以及文本特征$w^p_t\in \mathbb{R}^{d}$，最终的输出结果$p^{t}$通过公式1计算得出。老师模型的训练过程就是通过交叉熵损失的方法最小化预测概率与真实标签之间的距离来优化模型的参数。
% 接下来有意思的是我们的轻量级CLIP学生模型中的文本编码器分支直接复用老师模型中的文本特征，这样节约了学生模型中文本编码分支的训练成本。同时，相当于我们使用一个轻量级的CLIP视觉编码分支取代了大CLIP的视觉编码分支，却仍然能够保持具有竞争力的效果，这也减少了训练过程中计算资源的消耗。在学生模型的视觉编码分支中，我们的输入是没有标签的域数据集$D_{unlabeled}$ ，与老师模型的不同在于我们添加了可学习的项目层，得到的视觉特征$u^p_s\in \mathbb{R}^{d}$先与老师模型中的视觉特征$u^p_t\in \mathbb{R}^{d}$进行特征对齐，再经过我们提出的EduAttention模块，然后与预先存储好的老师模型中的文本特征进行融合，得到学生模型的标签，最后再通过KL散度的方法来拉近老师模型的标签以及学生模型的标签之间的距离进行知识蒸馏，在知识蒸馏以及学生模型视觉特征和老师模型视觉特征对齐的过程中优化学生模型的参数来提升学生模型的性能。

\subsection{ComKD-CLIP}

\noindent \textbf{IFAlign:}
The schematic of IFAlign is illustrated in Fig.~\ref{fig:2}(c). To make the image features extracted by the student model closely match those extracted by the teacher model, we align the mean and variance statistics for extracted features. The calculation process can be formulated as follows:

\begin{equation}
\begin{split}
\label{eq:stu-prompts}
&\bm{\mu}_{\text{s}}(u_{s};\bm{p}) = \frac{1}{N} \sum P(\bm{u}^p_{s}),\\
&\bm{\mu}_{\text{t}}(u_{t};\bm{p}) = \frac{1}{N} \sum \bm{u}^p_{t},
\end{split}
\end{equation}

\begin{equation}
\begin{split}
\label{eq:sigma-prompts}
&\bm{\sigma}^2_{s}(u_{s} ; \bm{p}) = \frac{1}{N} \sum \bigg(P(\bm{u}^p_{s}) - \bm{\mu}_{s}(u_{s} ; \bm{p})\bigg)^2,\\
&\bm{\sigma}^2_{t}(u_{t} ; \bm{p}) = \frac{1}{N} \sum \bigg(\bm{u}^p_{t} - \bm{\mu}_{t}(u_{t} ; \bm{p})\bigg)^2,
\end{split}
\end{equation}
where $\bm{\mu}_{\text{s}}(u_{\text{s}};\bm{p})$ and $\bm{\sigma}^2_{\text{s}}(u_{\text{s}} ; \bm{p})$ represent the means and variances of the image features extracted by the student model; $\bm{\mu}_{\text{t}}(u_{\text{t}};\bm{p})$ and $\bm{\sigma}^2_{\text{t}}(u_{\text{t}} ; \bm{p})$ correspond to those extracted by the teacher model. $\bm{u}^p_{\text{s}}$ and $\bm{u}^p_{\text{t}}$ denote the image features with prompts for the student and teacher models, respectively. The learnable projector $P(\cdot)$ within the student's image encoder branch is designed to adjust feature dimensions efficiently and cost-effectively, ensuring precise alignment. Following this, we use the $L_{1}$ loss to align the mean and variance of the image features extracted by student model with teacher model. This alignment can facilitate the student model to absorb the knowledge of how the teacher model extracts salient image features. The specific alignment process can be formulated as follows:

\begin{equation}
\begin{split}
\label{eq:align-loss_mean}
&\mathcal{L}_{\text{align\_mean}} = \| \bm{\mu}_{\text{s}}(u_{\text{s}};\bm{p}) - \bm{\mu}_{\text{t}}(u_{\text{t}};\bm{p}) \|_1 ,\\
&\mathcal{L}_{\text{align\_var}} = \| \bm{\sigma}^2_{s}(u_{s} ; \bm{p}) - \bm{\sigma}^2_{t}(u_{t} ; \bm{p}) \|_1 ,\\
&\mathcal{L}_{\text{align}} =  \mathcal{L}_{\text{align\_mean}} + \mathcal{L}_{\text{align\_var}},
\end{split}
\end{equation}
where $\mathcal{L}_{\text{align\_mean}}$ represents the difference between the mean values of image features extracted by the teacher model and the student model, ${L}_{\text{align\_var}}$ represents the difference between the variances values of image features extracted by the teacher model and the student model. Combining $\mathcal{L}_{\text{align\_mean}}$ and ${L}_{\text{align\_var}}$ as the alignment loss $\mathcal{L}_{\text{align}}$ allows the student model to fully absorb the knowledge of how the teacher model extracts image features.

\noindent \textbf{EduAttention:}
The schematic of EduAttention is illustrated in Fig.~\ref{fig:2}(e). In this module, an attention mechanism is leveraged to explore the cross-relationship between the image features extracted by student model and text features provided by teacher model, which can facilitate the student model to learn the nuanced strategies employed by the teacher model for integrating text-image feature. The specific calculation process can be formulated as follows:

\begin{equation}
\begin{split}
\label{eq:cls}
&\mathrm{\textit{Q}} = FC(\bm{w}^p_t), \ \ \ \ \mathrm{\textit{K}} = FC(\bm{u}^p_s), \ \ \ \ \mathrm{\textit{V}} = FC(\bm{u}^p_s), \\
&f_{att} = Softmax\left(\frac{\mathrm{\textit{Q}}\mathrm{\textit{K}}^T}{\sqrt{\mathrm{\textit{C}}}}\right) \cdot \mathrm{\textit{V}},\\
\end{split}
\end{equation}
where $\bm{u}^p_s$ represents the image features extracted by the student models, $\bm{w}^p_t$ represents the text features extracted by the teacher models, $f_{att}$ represents the cross-relationship between $\bm{u}^p_s$ and $\bm{w}^p_t$, $C$ is a hyperparameter, and $FC(\cdot)$ represents the fully connected layer.

To integrate the knowledge absorbed from the teacher model by IFAlign and EduAttention, we multiply  $f_{att}$ by a learnable parameter $\alpha$ and perform a element-wise sum operation with the extracted image features text $\bm{u}^p_t$ to final image features $f_e$. The specific calculation process can be formulated as follows:
\begin{equation}
\begin{split}
&f_e = \bm{u}^p_s + \mathrm{\alpha} \cdot f_{att},
\end{split}
\end{equation}
where $\alpha$ is initialized as 0 and gradually learns to assign more weight. 

\begin{table*}[!h]
    \centering
    \captionsetup{skip=5pt}
    \caption{We compare base-to-novel generalization capabilities with current state-of-the-art methods. Our \textbf{ComKD-CLIP} framework demonstrates exceptional generalization across 11 recognition datasets, employing the \textbf{ViT-B/16 image encoder} from the CLIP model. The symbol $\Delta$ represents the performance improvements relative to the previously established state-of-the-art, \textbf{PromptKD}.
}
    \renewcommand{\arraystretch}{1} % Adjust the row height
    % First row of subtables
     \vspace{5pt}
    \begin{minipage}{\textwidth}
        \centering
        \begin{subtable}[t]{0.32\linewidth}
            \centering
            \captionsetup{skip=2pt}
            \caption{Average over 11 datasets}
            \resizebox{\linewidth}{!}{
            \begin{tabular}{cccc}
                \hline
                ViT-B/16 & Base  & Novel & HM \\
                \hline
                CLIP & 69.34 & 74.22 & 71.70 \\
                CoOp & 82.69 & 63.22 & 71.66 \\
                CoCoOp & 80.47 & 71.69 & 75.83 \\
                MaPLe & 82.28 & 75.14 & 78.55 \\
                PromptSRC & 84.26 & 76.10 & 79.97 \\
                PromptKD &86.96 & 80.73 & 83.73 \\
                \hline
                \cellcolor{lightgray!30}ComKD-CLIP & \cellcolor{lightgray!30}87.37 & \cellcolor{lightgray!30}80.59 & \cellcolor{lightgray!30}83.84 \\
                $\Delta$ & \textcolor{customred}{\textbf{+0.41}} & -0.14 & \textcolor{customred}{\textbf{+0.11}} \\
                \hline
            \end{tabular}
            }
        \end{subtable} 
        \hfill
        \begin{subtable}[t]{0.32\linewidth}
            \centering
            \captionsetup{skip=2pt}
            \caption{Caltech101}
            \resizebox{\linewidth}{!}{
            \begin{tabular}{cccc}
                \hline
                ViT-B/16 & Base  & Novel & HM \\
                \hline
                CLIP & 96.84 & 94.00 & 95.40 \\
                CoOp & 98.00 & 89.81 & 93.73 \\
                CoCoOp & 97.96 & 93.81 & 95.84 \\
                MaPLe  & 97.74 & 94.36 & 96.02 \\
                PromptSRC & 98.10 & 94.03 & 96.02 \\
                PromptKD & 98.91 & 96.65 & 97.77 \\
                \hline
                \cellcolor{lightgray!30} ComKD-CLIP & \cellcolor{lightgray!30}99.23 & \cellcolor{lightgray!30}96.40 & \cellcolor{lightgray!30}97.79 \\
                $\Delta$ & \textcolor{customred}{\textbf{+0.32}} & -0.25 & \textcolor{customred}{\textbf{+0.02}} \\
                \hline
            \end{tabular}
            }
        \end{subtable}
        \hfill
        \begin{subtable}[t]{0.32\linewidth}
            \centering
            \captionsetup{skip=2pt}
            \caption{OxfordPets}
            \resizebox{\linewidth}{!}{
            \begin{tabular}{cccc}
                \hline
                ViT-B/16 & Base  & Novel & HM \\
                \hline
                CLIP & 91.17 & 97.26 & 94.12 \\
                CoOp & 93.67 & 95.29 & 94.47 \\
                CoCoOp & 95.20 & 97.69 & 96.43 \\
                MaPLe  & 95.43 & 97.76 & 96.58 \\
                PromptSRC & 95.33 & 97.30 & 96.30 \\
                PromptKD & 96.30 & 98.01 & 97.15 \\
                \hline
                \cellcolor{lightgray!30}ComKD-CLIP & \cellcolor{lightgray!30}96.76 & \cellcolor{lightgray!30}98.10 & \cellcolor{lightgray!30}97.43 \\
                $\Delta$ & \textcolor{customred}{\textbf{+0.46}} & \textcolor{customred}{\textbf{+0.09}} & \textcolor{customred}{\textbf{+0.28}} \\
                \hline
            \end{tabular}
            }
        \end{subtable}
    \end{minipage}
    \vspace{5pt}

    \begin{minipage}{\textwidth}
        \centering
        \begin{subtable}[t]{0.32\linewidth}
            \centering
            \captionsetup{skip=2pt}
            \caption{Flowers102}          
            \resizebox{\linewidth}{!}{
            \begin{tabular}{cccc}
                \hline
                ViT-B/16 & Base  & Novel & HM \\
                \hline
                CLIP & 72.08 & 77.80 & 74.83 \\
                CoOp & 97.60 & 59.67 & 74.06 \\
                CoCoOp & 94.87 & 71.75 & 81.71 \\
                MaPLe & 95.92 & 72.46 & 82.56 \\
                PromptSRC & 98.07 & 76.50 & 85.95 \\
                PromptKD & 99.42 & 82.62 & 90.24 \\
                \hline
                \cellcolor{lightgray!30}ComKD-CLIP & \cellcolor{lightgray!30}99.53 & \cellcolor{lightgray!30}82.62 & \cellcolor{lightgray!30}90.29 \\
                $\Delta$ & \textcolor{customred}{\textbf{+0.11}} & +0.00 & \textcolor{customred}{\textbf{+0.05}} \\
                \hline
            \end{tabular}
            }
        \end{subtable}
        \hfill
        \begin{subtable}[t]{0.32\linewidth}
            \centering
            \captionsetup{skip=2pt}
            \caption{Food101}
            \resizebox{\linewidth}{!}{
            \begin{tabular}{cccc}
                \hline
                ViT-B/16 & Base  & Novel & HM \\
                \hline
                CLIP & 90.10 & 91.22 & 90.66 \\
                CoOp & 88.33 & 82.26 & 85.19 \\
                CoCoOp & 90.70 & 91.29 & 90.99 \\
                MaPLe & 90.71 & 92.05 & 91.38 \\
                PromptSRC & 90.67 & 91.53 & 91.10 \\
                PromptKD & 92.43 & 93.68 & 93.05 \\
                \hline
                \cellcolor{lightgray!30}ComKD-CLIP & \cellcolor{lightgray!30}92.78 & \cellcolor{lightgray!30}93.91 & \cellcolor{lightgray!30}93.34 \\
                $\Delta$ & \textcolor{customred}{\textbf{+0.35}} & \textcolor{customred}{\textbf{+0.23}} & \textcolor{customred}{\textbf{+0.29}} \\
                \hline
            \end{tabular}
            }
        \end{subtable}
        \hfill
        \begin{subtable}[t]{0.32\linewidth}
            \centering
            \captionsetup{skip=2pt}
            \caption{FGVCAircraft}
            \resizebox{\linewidth}{!}{
            \begin{tabular}{cccc}
                \hline
                ViT-B/16 & Base  & Novel & HM \\
                \hline
                CLIP & 27.19 & 36.29 & 31.09 \\
                CoOp & 40.44 & 22.30 & 28.75 \\
                CoCoOp & 33.41 & 23.71 & 27.74 \\
                MaPLe & 37.44 & 35.61 & 36.50 \\
                PromptSRC & 42.73 & 37.87 & 40.15 \\
                PromptKD & 49.12 & 41.81 & 45.17 \\
                \hline
                \cellcolor{lightgray!30}ComKD-CLIP & \cellcolor{lightgray!30}51.80 & \cellcolor{lightgray!30}43.37 & \cellcolor{lightgray!30}47.21  \\
                $\Delta$ & \textcolor{customred}{\textbf{+2.68}} & \textcolor{customred}{\textbf{+1.56}} & \textcolor{customred}{\textbf{+2.04}} \\
                \hline
            \end{tabular}
            }
        \end{subtable}
    \end{minipage}
    \vspace{5pt} 
    
    \begin{minipage}{\textwidth}
        \centering    
        \begin{subtable}[t]{0.32\linewidth}
            \centering
            \captionsetup{skip=2pt}
            \caption{SUN397}
            \resizebox{\linewidth}{!}{
            \begin{tabular}{cccc}
                \hline
                ViT-B/16 & Base  & Novel & HM \\
                \hline
                CLIP & 69.36 & 75.35 & 72.23 \\
                CoOp & 80.60 & 65.89 & 72.51 \\
                CoCoOp & 79.74 & 76.86 & 78.27 \\
                MaPLe & 80.82 & 78.70 & 79.75 \\
                PromptSRC & 82.67 & 78.47 & 80.52 \\
                PromptKD & 83.69 & 81.54 & 82.60 \\
                \hline
                \cellcolor{lightgray!30}ComKD-CLIP & \cellcolor{lightgray!30}84.19 & \cellcolor{lightgray!30}81.70 & \cellcolor{lightgray!30}82.93 \\
                $\Delta$ & \textcolor{customred}{\textbf{+0.50}} & \textcolor{customred}{\textbf{+0.16}} & \textcolor{customred}{\textbf{+0.33}} \\
                \hline
            \end{tabular}
            }
        \end{subtable}
        \hfill
        \begin{subtable}[t]{0.32\linewidth}
            \centering
            \captionsetup{skip=2pt}
            \caption{DTD}
            \resizebox{\linewidth}{!}{
            \begin{tabular}{cccc}
                \hline
                ViT-B/16 & Base  & Novel & HM \\
                \hline
                CLIP & 53.24 & 59.90 & 56.37 \\
                CoOp & 79.44 & 41.18 & 54.24 \\
                CoCoOp & 77.01 & 56.00 & 64.85 \\
                MaPLe & 80.36 & 59.18 & 68.16 \\
                PromptSRC & 83.37 & 62.97 & 71.75 \\
                PromptKD & 85.84 & 71.37 & 77.94 \\
                \hline
                \cellcolor{lightgray!30}ComKD-CLIP & \cellcolor{lightgray!30}86.46 & \cellcolor{lightgray!30}70.89 & \cellcolor{lightgray!30}77.90 \\
                $\Delta$ & \textcolor{customred}{\textbf{+0.62}} & -0.48 & -0.04 \\
                \hline
            \end{tabular}
            }
        \end{subtable}
        \hfill
        \begin{subtable}[t]{0.32\linewidth}
            \centering
            \captionsetup{skip=2pt}
            \caption{EuroSAT}
            \resizebox{\linewidth}{!}{
            \begin{tabular}{cccc}
                \hline
                ViT-B/16 & Base  & Novel & HM \\
                \hline
                CLIP & 56.48 & 64.05 & 60.03 \\
                CoOp & 92.19 & 54.74 & 68.69 \\
                CoCoOp & 87.49 & 60.04 & 71.21 \\
                MaPLe & 94.07 & 73.23 & 82.35 \\
                PromptSRC & 92.90 & 73.90 & 82.32 \\
                PromptKD & 97.54 & 82.08 & 89.14 \\
                \hline
                \cellcolor{lightgray!30}ComKD-CLIP & \cellcolor{lightgray!30}98.17 & \cellcolor{lightgray!30}81.79 & \cellcolor{lightgray!30}89.23 \\
                $\Delta$ & \textcolor{customred}{\textbf{+0.63}} & -0.29 & \textcolor{customred}{\textbf{+0.09}} \\
                \hline
            \end{tabular}
            }
        \end{subtable}
    \end{minipage}
    \vspace{5pt} 
    
    \begin{minipage}{\textwidth}
        \centering
        \begin{subtable}[t]{0.32\linewidth}
            \centering
            \captionsetup{skip=2pt}
            \caption{UCF101}
            \resizebox{\linewidth}{!}{
            \begin{tabular}{cccc}
                \hline
                ViT-B/16 & Base  & Novel & HM \\
                \hline
                CLIP & 70.53 & 77.50 & 73.85 \\
                CoOp & 84.69 & 56.05 & 67.46 \\
                CoCoOp & 82.33 & 73.45 & 77.64 \\
                MaPLe & 83.00 & 78.66 & 80.77 \\
                PromptSRC & 87.10 & 78.80 & 82.74 \\
                PromptKD & 89.71 & 82.27 & 86.10 \\
                \hline
                \cellcolor{lightgray!30}ComKD-CLIP & \cellcolor{lightgray!30}90.28 & \cellcolor{lightgray!30}81.45 & \cellcolor{lightgray!30}85.64 \\
                $\Delta$ & \textcolor{customred}{\textbf{+0.57}} & -0.82 & -0.46 \\
                \hline
            \end{tabular}
            }
        \end{subtable}
        \hfill
        \begin{subtable}[t]{0.32\linewidth}
            \centering
            \captionsetup{skip=2pt}
            \caption{StanfordCars}
            \resizebox{\linewidth}{!}{
            \begin{tabular}{cccc}
                \hline
                ViT-B/16 & Base  & Novel & HM \\
                \hline
                CLIP & 63.37 & 74.89 & 68.65 \\
                CoOp & 78.12 & 60.40 & 68.13 \\
                CoCoOp & 70.49 & 73.59 & 72.01 \\
                MaPLe & 72.94 & 74.00 & 73.47 \\
                PromptSRC & 78.27 & 74.97 & 76.58 \\
                PromptKD & 82.80 & 83.37 & 83.13 \\
                \hline
                \cellcolor{lightgray!30}ComKD-CLIP & \cellcolor{lightgray!30}85.33 & \cellcolor{lightgray!30}85.19 & \cellcolor{lightgray!30}85.26 \\
                $\Delta$ & \textcolor{customred}{\textbf{+2.53}} & \textcolor{customred}{\textbf{+1.82}} & \textcolor{customred}{\textbf{+2.13}} \\
                \hline
            \end{tabular}
            }
        \end{subtable}
        \hfill
        \begin{subtable}[t]{0.32\linewidth}
            \centering
            \captionsetup{skip=2pt}
            \caption{ImageNet}
            \resizebox{\linewidth}{!}{
            \begin{tabular}{cccc}
                \hline
                ViT-B/16 & Base  & Novel & HM \\
                \hline
                CLIP & 72.43 & 68.14 & 70.22 \\
                CoOp & 76.47 & 67.88 & 71.92 \\
                CoCoOp & 75.98 & 70.43 & 73.10 \\
                MaPLe  & 76.66 & 70.54 & 73.47 \\
                PromptSRC & 77.60 & 70.73 & 74.01 \\
                PromptKD & 80.83 & 74.66 & 77.62 \\
                \hline
                \cellcolor{lightgray!30}ComKD-CLIP & \cellcolor{lightgray!30}76.53 & \cellcolor{lightgray!30}71.02 & \cellcolor{lightgray!30}73.67 \\
                $\Delta$ & -4.30 & -3.64 & -3.95 \\
                \hline
            \end{tabular}
            }
        \end{subtable}
    \end{minipage}
    \label{table:base2novel}
    \vspace{-10pt}
\end{table*}

\noindent \textbf{Distilled Knowledge Refinement:}
After the student model absorbs the knowledge of how the teacher model extracts image features and combines text-image features, we try to refine the absorbed knowledge based on the feature fusion results of the teacher model. As depicted in Fig.~\ref{fig:2}, we utilize the KL divergence  to minimize the discrepancy between the feature distribution produced by the teacher and student models. The specific process can be formulated as follows:

\begin{equation}
\begin{split}
&\mathcal{L}_{stu}=\mathcal{L}_{\text{kd}}(q^{t}, q^{s}, \tau), \\
\end{split}
\end{equation}
where $q^{t}$ and $q^{s}$ represent the logits predicted by the teacher and student models, respectively, which is calculated by the corresponding image features and text features using Eq.~\ref{equation:output_prob}. $\tau$ is the temperature parameter, which is used to adjust the smoothness of the probability distribution.

Finally, we combine the student model's alignment loss $\mathcal{L}_{\text{stu}}$ with the feature distribution loss $\mathcal{L}_{\text{align}}$ as the final loss function to train the parameters of the small CLIP model, the specific loss formula is shown as follows:

\begin{equation}
\label{eq:final}
    \mathcal{L}_{\text{final}} = \mathcal{L}_{\text{stu}}   + \mathcal{L}_{\text{align}}.
\end{equation}

\section{Experiments}

\subsection{Settings}
\textbf{Datasets:} In this study, we adopt the methodologies from PromptSRC~\cite{khattak2023self} and PromptKD~\cite{li2024promptkd} to assess the generalization from a base to novel classes, cross-dataset evaluation. We utilize 11 diverse image recognition datasets encompassing a range of tasks such as:

- Generic object recognition with ImageNet~\cite{deng2009imagenet} and Caltech101~\cite{fei2004learning}.

- Fine-grained classification using OxfordPets~\cite{parkhi2012cats}, StanfordCars~\cite{krause20133d}, Flowers102~\cite{nilsback2008automated}, Food101~\cite{bossard2014food}, and FGVCAircraft~\cite{maji2013fine}.

- Scene recognition with SUN397~\cite{xiao2010sun}.

- Action recognition from UCF101~\cite{soomro2012ucf101}.

- Texture classification via DTD~\cite{cimpoi2014describing}.

- Satellite imagery with EuroSAT~\cite{helber2019eurosat}.

For the domain generalization benchmark, ImageNet~\cite{deng2009imagenet} serves as the source dataset, with ImageNetA~\cite{hendrycks2021natural}, ImageNet-R~\cite{hendrycks2021many}, ImageNet-Sketch~\cite{wang2019learning}, and ImageNetV2~\cite{recht2019imagenet} as the out-of-distribution test datasets.

\noindent \textbf{Implementation Details:} 
We employ the ViT-L/14 CLIP model as the teacher model and the ViT-B/16 CLIP model as the student model for our ComKD-CLIP framework. Following the PromptKD configuration, we set the prompt depth to 9, with both vision and language prompt lengths fixed at 4. Optimization is carried out using Stochastic Gradient Descent (SGD) with the temperature hyperparameter $\tau$ set to its default value of 1. Initial text prompts for the first layer are generated using embeddings of the phrase "a photo of a \{classname\}". We report the base and novel class accuracies along with their Harmonic Mean (HM), averaged over three runs. All experiments are conducted on a single Nvidia A100 GPU.

\subsection{Base-to-novel Generalization}
Following~\cite{zhou2022conditional}~\cite{khattak2023maple}~\cite{khattak2023self}~\cite{li2024promptkd}, we divide the training and testing datasets into base and novel classes. Our teacher model, pre-trained via the PromptSRC approach~\cite{khattak2023self}, guides the training of student model using an unlabeled set. Post-distillation, we assess the students' performance on both class types against the test set, serving as a measure of methodological generalization within the dataset.
As shown in Table~\ref{table:base2novel}, we compare the performance of our proposed ComKD-CLIP with recent state-of-the-art methods including CLIP~\cite{radford2021learning}, CoOp~\cite{zhou2022learning}, CoCoOp~\cite{zhou2022conditional}, MaPLe~\cite{khattak2023maple}, PromptSRC~\cite{khattak2023self}, PromptKD~\cite{li2024promptkd} on 11 recognition datasets. In comparison with these state-of-the-art works, ComKD-CLIP shows highly competitive results in all 11 datasets.

\subsection{Cross-dataset Evaluation}
Similar to PromptKD~\cite{li2024promptkd}, our teacher model undergoes pre-training on ImageNet~\cite{deng2009imagenet}. Subsequently, we utilize the training set of unlabeled target datasets to train the student model. Their performance is then evaluated on the test set post-training, with no data-specific fine-tuning applied. We compare our cross-dataset performance with previous methods  in Table~\ref{table:cross_dataset}, our proposed ComKD-CLIP outperforms some state-of-the-art methods on 8 out of 10 datasets, achieving an average improvement of 0.74\% over previous methods.

\subsection{Domain Generalization Experiments}
We train a source model on the ImageNet~\cite{deng2009imagenet} dataset and subsequently evaluate its robustness across various out-of-distribution datasets to assess performance under domain shifts. This method helps us explore the model's adaptability to different and unexpected environments, thereby identifying its strengths and potential vulnerabilities in practical applications. We summarize the results of ComKD-CLIP and compare with previous methods on out-of-distribution datasets in Table ~\ref{table:ood}. The results indicate that ComKD-CLIP outperforms some state-of-the-art methods on the source datasets ImageNetV2, ImageNet-Sketch, and ImageNetA, with higher average performance as well. This demonstrates that ComKD-CLIP has stronger generalization capabilities on datasets with domain shifts. 
\begin{table*}[!ht]
    \centering
    \captionsetup{skip=5pt}
    \caption{The performance of \textbf{ComKD-CLIP} with some state-of-the-art mothods in a cross-dataset benchmark. Utilizing our pipeline, we conduct unsupervised Aligned distillation with unlabeled domain data in a transductive setting. The source model is pretrained on ImageNet. "ZSL" indicates that the evaluation is conducted in a Zero-Shot Learning setting. \textbf{ComKD-CLIP} outperforms on 8 out of 10 datasets.}
    \renewcommand{\arraystretch}{1} % Adjust the row height
    \resizebox{\linewidth}{!}
    {
        \begin{tabular}{ccccccccccccc}
        \hline\noalign{\smallskip}
        ~   & ~         & \multicolumn{10}{c}{\textbf{Target Dataset}} \\
        \cmidrule(lr){3-13} % Add a horizontal line under Target Dataset
        \multirow{2}*{ZSL} & \multirow{2}*{ViT-B/16} & Caltech & Oxford  & Flowers & \multirow{2}*{Food101} & FGVC & \multirow{2}*{SUN397} & \multirow{2}*{DTD} & Euro & \multirow{2}*{UCF101} & Stanford & \multirow{2}*{Avg.} \\
        ~   &          & 101     & Pets      & 102     & ~  & Aircraft & ~ & ~     & SAT & ~ & Cars & ~    \\
        \cmidrule(lr){1-13} % Add a full horizontal line
        \noalign{\smallskip}
        ~ & CoOp       & 93.70 & 89.14 & 68.71 & 85.30 & 18.47 & 64.15 & 41.92 & 46.39 & 66.55 & 64.51& 63.88 \\
        In- & CoCoOp  & 94.43 & 90.14 & 71.88 & 86.06 & 22.94 & 67.36 & 45.73 & 45.37 & 68.21 &65.32 & 65.74 \\
        ductive & MaPLe & 93.53 & 90.49 & 72.23 & 86.20 & 24.74 & 67.01 & 46.49 & 48.06 & 68.69 & 65.57& 66.30 \\
        ~ & PromptSRC   & 93.60 & 90.25 & 70.25 & 86.15 & 23.90 & 67.10 & 46.87 & 45.50 & 68.75 &65.70 & 65.81 \\
        \cmidrule(lr){1-13} % Add a full horizontal line
        \noalign{\smallskip}
         Trans-& PromptKD  & 93.61 & \textbf{91.59} & \textbf{75.33} & 88.84 & 26.24 & 68.57 & 55.08 & 63.74 & 76.39 &73.93 & 71.33 \\
         & ComKD-CLIP &  \textbf{94.56}  & 91.36 & 75.07  &  \textbf{89.21}  & \textbf{27.54} & \textbf{69.82} & \textbf{57.09} &  \textbf{64.27} & \textbf{77.16} & \textbf{74.59} & \textbf{72.07}\\
        ductive & $\Delta$ & \textcolor{customred}{\textbf{+0.95}} & -0.23 & -0.26 & \textcolor{customred}{\textbf{+0.37}} & \textcolor{customred}{\textbf{+1.30}} & \textcolor{customred}{\textbf{+1.25}} & \textcolor{customred}{\textbf{+2.01}} & \textcolor{customred}{\textbf{+0.53}} & \textcolor{customred}{\textbf{+0.77}} &\textcolor{customred}{\textbf{+0.66}} & \textcolor{customred}{\textbf{+0.74}} \\
        \hline
        \end{tabular}
    }
    
    \label{table:cross_dataset}
    \vspace{-10pt}
\end{table*}

\begin{table}[!h]
    \centering
    \captionsetup{skip=5pt}
    \caption{The performance of ComKD-CLIP within domain generalization contexts. The method is trained on ImageNet and subsequently evaluated on datasets with domain shifts.
    }
    \renewcommand{\arraystretch}{1} % Adjust the row height
    \resizebox{\linewidth}{!}{
    \begin{tabular}{cccccc}
        \hline\noalign{\smallskip}
         ~          & \multicolumn{5}{c}{\textbf{Target Dataset}} \\
        \cmidrule(lr){2-6} % Add a horizontal line under Target Dataset
        ViT-B/16  & -V2  & -S & -A & -R & Avg. \\
        \cmidrule(lr){1-6} % Add a full horizontal line
        \noalign{\smallskip}
        CLIP & 60.83 & 46.15 & 47.77 & 73.96 & 57.18 \\
        CoOp & 64.20 & 47.99 & 49.71 & 75.21 & 59.28  \\
        CoCoOp &  64.07 & 48.75 & 50.63 & 76.18 & 59.91  \\
        MapLe &  64.07 & 49.15 & 50.90 & 76.98 & 60.27 \\
        PromptSRC &  64.35 & 49.55 & 50.90 & 77.80 & 60.65  \\
        PromptKD  &  69.77 & 58.72 & 70.36 & \textbf{87.01} & 71.47  \\
        \cmidrule(lr){1-6} % Add a full horizontal line
        \noalign{\smallskip}
        ComKD-CLIP &  \textbf{70.95} & \textbf{59.92} & \textbf{74.15} & 86.08 & \textbf{72.78} \\
        $\Delta$ &  \textcolor{customred}{\textbf{+1.18}} & \textcolor{customred}{\textbf{+1.2}} & \textcolor{customred}{\textbf{+3.79}} & -0.93 & \textcolor{customred}{\textbf{+1.31}}\\
        \hline
    \end{tabular} 
    }
        
    \label{table:ood}
    \vspace{-10pt}
\end{table}

\section{Ablation Study}

\subsection{The effectiveness of IFAlign \& EduAttention}

To begin with, we try to explore the effectiveness of IFAlign module and  EduAttention module. Specifically, we remove the IFAlign module and EduAttention module from ComKD-CLIP respectively and test the corresponding experimental results. As shown in Table~\ref{tab: align & eduattention}, removing IFAlign module will greatly reduce the performance of ComKD-CLIP. Similarly, removing the EduAttention module also decrease the model's performance. It is worth noting that when both the IFAlign module and the EduAttention module are removed, ComKD-CLIP has the worst performance. These experimental results fully prove that IFAlign module and EduAttention module can 
effectively prompt the student model to absorb the teacher model's knowledge on how to extract image features and integrate text-image feature.

\begin{table}[!ht]
    \centering
    \captionsetup{skip=5pt}
    \caption{Ablation study of IFAlign module and EduAttention module on SUN397.}
    \renewcommand{\arraystretch}{1.1} % Adjust the row height
    \resizebox{\linewidth}{!}{
    \begin{tabular}{lccc}
        \hline
        Methods & Base  & Novel & HM \\
        \hline
        ComKD-CLIP w/o IFAlign & 83.89 & 81.68 & 82.77 \\
        ComKD-CLIP w/o EduAttention & 83.83 & \textbf{81.70} & 82.75 \\
        ComKD-CLIP w/o IFAlign & \multirow{2}{*}{83.69} & \multirow{2}{*}{81.54} & \multirow{2}{*}{82.60} \\
        w/o EduAttention & &  &  \\
        \hline
        \cellcolor{lightgray!30}ComKD-CLIP (Full) & \cellcolor{lightgray!30}\textbf{84.19} & \cellcolor{lightgray!30}\textbf{81.70} & \cellcolor{lightgray!30}\textbf{82.93} \\
        \hline
    \end{tabular}
    }
    
    \label{tab: align & eduattention}
    \vspace{-10pt}
\end{table}

Subsequently, we try to use different alignment methods in IFAlign to find the best alignment strategy. We respectively align image features by mean ($\mathcal{L}_{\text{align\_mean}}$), variance ($\mathcal{L}_{\text{align\_var}}$), and mean add variance ($\mathcal{L}_{\text{align}}$). The experimental results are shown in  Table~\ref{tab:align}. It can be clearly seen that ComKD-CLIP has the best performance when we use the mean add variance ($\mathcal{L}_{\text{align}}$) to align image features in IFAlign. This mainly because that mean add variance can jointly promote the alignment from the perspectives of center position and discreteness, thereby helping the student model to absorb the teacher model's knowledge on how to extract image features.

\begin{table}[!ht]
    \centering
    \captionsetup{skip=5pt}
    \caption{Ablation study of alignment strategy used in IFAlign on SUN397.}
    \renewcommand{\arraystretch}{1.3} % Adjust the row height
    \resizebox{0.67\linewidth}{!}{
    \begin{tabular}{lccc}
        \hline
        Methods & Base  & Novel & HM \\
        \hline
         $\mathcal{L}_\text{align\_var}$ & 83.88 & 81.21 & 82.52 \\
        $\mathcal{L}_\text{align\_mean}$ & 83.84 & \textbf{81.93} & 82.87 \\
        \hline
        \cellcolor{lightgray!30}$\mathcal{L}_\text{align}$ & \cellcolor{lightgray!30}\textbf{84.19} & \cellcolor{lightgray!30}81.70 & \cellcolor{lightgray!30}\textbf{82.93} \\
        \hline
    \end{tabular}
    }
    \label{tab:align}
    \vspace{-10pt}
\end{table}

\subsection{Knowledge Refinement Method}

Finally, we try to explore the best strategy to refine the knowledge distilled by IFAlign module and EduAttention module.
Specifically, we use KL divergence, L1 and MSE methods to refine the distilled knowledge. The corresponding performance of ComKD-CLIP is shown in Table~\ref{tab:kl method}. It can be seen that KL divergence can better refine the distilled knowledge and make ComKD-CLIP perform best. This may be because KL divergence can make the logits of the student model better approximate the logits of the teacher model, thus more effectively refining the knowledge distilled by the student model.

\begin{table}[!ht]
    \centering
    \captionsetup{skip=5pt}
    \caption{Comparison between different distillation methods in SUN397.}
    \renewcommand{\arraystretch}{1.3} % Adjust the row height
    \resizebox{0.65\linewidth}{!}{
    \begin{tabular}{lccc}
        \hline
        KD Loss & Base  & Novel & HM \\
        \hline
        L1 & 66.16 & 58.84 & 62.29 \\
        MSE & 67.79 & 61.93 & 64.73 \\
        \hline
        \cellcolor{lightgray!30}KL & \cellcolor{lightgray!30}\textbf{84.19} & \cellcolor{lightgray!30}\textbf{81.70} & \cellcolor{lightgray!30}\textbf{82.93} \\
        \hline
    \end{tabular}
    }
    
    \label{tab:kl method}
    \vspace{-10pt}
\end{table}

%在这里，我们设计实验来证明我们选取的知识蒸馏方法的有效性，我们将KL散度方法用L1以及MSE方法进行替换，其他设置保持一致，从表6中可以看出，我们所采用的KL散度知识蒸馏方法效果更好，并且远远高于L1方法以及MSE方法这也证明知识蒸馏方法对提升学生模型的性能具有重要影响。

\section{Conclusion}

In this study, we present ComKD-CLIP, an advanced knowledge distillation framework designed to comprehensively distill the knowledge from
a large teacher CLIP model to a smaller student model.
This process maintains the comparable performance while substantially reducing the model's parameter. ComKD-CLIP innovatively employs IFAlign and EduAttention to effectively distill the intricate knowledge embedded within the teacher CLIP model during the text-image feature fusion process. Futhermore, ComKD-CLIP can refine the distilled knowledge by leveraging the feature fusion results of the teacher model, ensuring that the smaller student model can accurately absorb the knowledge from teacher model. Extensive experiments across 11 datasets have unequivocally demonstrated the superior performance of ComKD-CLIP. The proposed method significantly bolsters the capabilities of smaller CLIP models in resource-constrained environments, thereby broadening the practical utility of the CLIP technology.

\section{Acknowledgments}

This research was supported by the Natural Science Basic Research Program of Shaanxi (2024JC-YBMS-468) and the China National Key R\&D Program (2022YFC2808003).

% \section{References}

\bibliography{aaai25}
\end{document}